\documentclass[letterpaper, 10 pt, conference]{ieeeconf}   

\IEEEoverridecommandlockouts
\usepackage{cite}

\usepackage{amsmath}
\usepackage{todonotes}
\newtheorem{remark}{Remark}
\usepackage{booktabs}
\usepackage{multirow}

\usepackage[linesnumbered, ruled, vlined]{algorithm2e}

\newcommand{\prob}{{\rm Pr}}
\newcommand{\FK}{\boldsymbol{\rm FK}}

\newcommand{\LF}{\rm FR}
\newcommand{\RF}{\rm FL}
\newcommand{\LR}{\rm RR}
\newcommand{\RR}{\rm RL}

\newcommand{\meas}[1]{\bar{#1}}
\newcommand{\est}[1]{\hat{#1}}
\newcommand{\pseudoMeas}[1]{\hat{#1}}

\newcommand{\bodyframe}{{\rm bf}}
\newcommand{\worldframe}{{\rm wf}}

\newcommand{\eulerAngles}{\boldsymbol{\Theta}}
\newcommand{\estEulerAngles}{\est{\eulerAngles}}
\newcommand{\measEulerAngles}{\meas{\eulerAngles}}

\newcommand{\inertiaBF}{\boldsymbol{\rm I}^{\bodyframe}}

\newcommand{\rotationsSymbol}{\omega}
\newcommand{\rotationsWF}{\boldsymbol{\rotationsSymbol}^{\worldframe}}

\newcommand{\measRotationsBF}{\meas{\boldsymbol{\rotationsSymbol}}^{\bodyframe}}

\newcommand{\legJointPosSymbol}{q}
\newcommand{\estLegJointPos}{\est{\boldsymbol{\legJointPosSymbol}}}
\newcommand{\measLegJointPos}{\meas{\boldsymbol{\legJointPosSymbol}}}
\newcommand{\estLegJointVel}{\est{\dot{\boldsymbol{\legJointPosSymbol}}}}
\newcommand{\measLegJointVel}{\meas{\dot{\boldsymbol{\legJointPosSymbol}}}}

\newcommand{\measAccelerationsBF}{\meas{\boldsymbol{a}}^\bodyframe}

\newcommand{\RotBtoW}{\boldsymbol{R}_{\bodyframe\rightarrow \worldframe}}
\newcommand{\RotWtoB}{\boldsymbol{R}_{\worldframe\rightarrow \bodyframe}}

\newcommand{\I}{\boldsymbol I}
\newcommand{\0}{\boldsymbol 0}

\newcommand{\Jac}{\boldsymbol{J}}


\title{\LARGE \bf
Simultaneous State Estimation and Contact Detection for Legged Robots by Multiple-Model Kalman Filtering
}

\author{Marcel Menner and Karl Berntorp%
\thanks{This research was not funded by any government agency.}
\thanks{Mitsubishi Electric Research Laboratories (MERL), 201 Broadway, Cambridge, MA, 02139, USA (e-mail: \{menner,karl.o.berntorp\}@ieee.org).}%
}

\begin{document}

\maketitle
\thispagestyle{empty}
\pagestyle{empty}

\begin{abstract}
This paper proposes an algorithm for combined contact detection and state estimation for legged robots. The proposed algorithm models the robot's movement as a switched system, in which different modes relate to different feet being in contact with the ground. The key element in the proposed algorithm is an interacting multiple-model Kalman filter, which identifies the currently-active mode defining contacts, while estimating the state. The rationale for the proposed estimation framework is that contacts (and contact forces) impact the robot's state and vice versa. This paper presents validation studies with a quadruped using (i) the high-fidelity simulator Gazebo for a comparison with ground truth values and a baseline estimator, and (ii) hardware experiments with the Unitree A1 robot. The simulation study shows that the proposed algorithm outperforms the baseline estimator, which does not simultaneous detect contacts. The hardware experiments showcase the applicability of the proposed algorithm and highlights the ability to detect contacts. 
\end{abstract}

\section{Introduction}

Mobile robots are useful for a variety of tasks to increase automation or provide assistance, e.g., robots have been employed for warehouse logistics, search and rescue, load carrying, etc.
Wheeled robots have proven to be effective in more structured environments such as warehouses.
On the other hand, mobile robots that move by using legs provide a higher level of versatility and the potential to maneuver in more unstructured environments.
E.g., legged robots have the potential to maneuver uneven terrain, stairs, rubble, and the like, which is particularly important for search and rescue missions.
However, this higher level of versatility comes at the expense of greater difficulty to achieve accurate and safe movements, i.e., the legs have to be moved to simultaneously traverse an area, stabilize the robot, and avoid unwanted contacts with the environment.
To fully reap the benefits of a legged robot's versatility in unstructured environments, fast control and estimation algorithms are needed that target such environments.

This paper focuses on state estimation and contact detection. 
One possible path to achieving accurate state estimation for control is to introduce additional sensors in the robot's design such as force sensors in the robot's feet, a camera, a light detection and ranging (LiDAR), etc.
However, this path makes a robot more expensive and more sensitive to failures in the additional sensors.
This paper presents a software-based solution with an algorithm for state estimation that only uses inertial measurement unit (IMU) sensor measurements (Euler angles, rotations, linear accelerations) and motor measurements (joint angles, joint accelerations, motor torques). 
Such sensor measurements are commonly available on robotic platforms such as legged robots.

\subsection{Algorithm Realization and Contributions}

More detailed, this paper proposes an algorithm for simultaneous contact detection and state estimation for legged robots, which is based on a physics-based motion model to reason about the contact forces' impact on the robot.
The proposed algorithm uses an interacting multiple-model Kalman filter (IMM-KF), in which the robot's movement are modeled as a switched system.
The switched system relates the multiple models to different feet being in contact with the ground.
The IMM-KF identifies the currently-active mode defining contacts, while estimating the state.
The proposed algorithm is validated using hardware experiments with the Unitree A1 and the high-fidelity simulator Gazebo, which is a physics simulation provided by the robot manufacturer.
The high-fidelity simulator is used to compare the proposed algorithm with ground truth values and a baseline estimator.
The hardware experiments showcase the applicability of the proposed algorithm and highlights the ability to detect intentional and unintentional contacts.



\subsection{Related Work}

In~\cite{bloesch2013state2, bloesch2013state, bloesch2017state}, a stochastic filtering approach with outlier rejection is presented, which fuses leg kinematics with inertial measurements. 
In~\cite{camurri2017probabilistic}, a probabilistic framework for detecting contacts is presented, which are subsequently used to improve the state estimate of the robot. 
In~\cite{bledt2018contact}, a disturbance observer-based approach is proposed for contact detection with an event-based finite state machine. 
In~\cite{hartley2018legged}, a state estimator that uses a factor-graph framework is proposed that considers the robot's kinematic model as well as its contact with the environment.  
A factor graph optimization method is presented in~\cite{wisth2019robust} for state estimation fusing inertial navigation, leg odometry and visual odometry. 
In~\cite{hartley2018contact, hartley2020contact}, a contact-aided invariant extended Kalman filter is developed using the theory of Lie groups and invariant observer design.  
The work in~\cite{camurri2020pronto} fuses IMU and leg odometry sensing for pose and velocity estimation by relying on force/torque sensors at the feet using the Schmitt trigger mechanism for contact classification.
In~\cite{fahmi2021state}, IMU measurements are fused with leg odometry and studies the impact of soft terrain on state estimation.
In~\cite{teng2021legged}, a state estimator for slippery environments is proposed, which fuses inertial and velocity measurements from a tracking camera and leg kinematics; \cite{kim2021legged} presents a state estimation algorithm based on the Gauss-Newton algorithm and the Schur Complement;~\cite{lin2021legged} proposes a contact-aided invariant extended Kalman filter, in which a trained neural network estimates contact events on different terrains;~\cite{kim2022step} uses body speed measurements obtained from a stereo camera without the need for a non-slip assumption; and~\cite{buchanan2022learning} proposes a state estimator for legged robots based on learned displacement measurement using convolutional neural networks.
While current solutions often rely on a camera, which is susceptible to external conditions such as environmental/lighting, or the availability of contact sensors, the proposed state estimation algorithm only uses IMU measurements for simultaneous state estimation and contact detection. 
The proposed algorithm uses an IMM-KF to reason about likelihoods of contacts while estimating the state and is executable in real time.


\section{Preliminaries}
\label{sec:prelim}

\subsection{Notation} 
This paper uses boldface $\boldsymbol{x}$ and boldface $\boldsymbol{X}$ to indicate a vector and matrix, respectively.
This paper uses parenthesis, $\boldsymbol{x}(t)$, and sub-scripts, $\boldsymbol{x}_t$, to distinguish the state in continuous time and discrete time $t$, respectively.
Let $\prob(\mathcal{X})$ be the probability of an event $\mathcal{X}$, e.g., $\prob(\delta=1)$ is the probability that $\delta=1$.
$\boldsymbol{P}_{\rm skew}(\boldsymbol{p})\in\mathbf{R}^{3\times 3}$ is the skew matrix of a vector $\boldsymbol{p}\in\mathbf{R}^3$ such that $\boldsymbol{P}_{\rm skew}(\boldsymbol{p})\boldsymbol{f} = \boldsymbol{p} \times \boldsymbol{f}$ with the cross product $\times$.
The notation $\boldsymbol{x}\sim \mathcal{N}(\boldsymbol{\mu},\boldsymbol{\Sigma})$ means $\boldsymbol{x}$ sampled from a normal distribution, $\mathcal{N}(\boldsymbol{\mu},\boldsymbol{\Sigma})$ with mean~$\boldsymbol{\mu}$ and covariance~$\boldsymbol{\Sigma}$.

Throughout, this paper uses super-scripts ${(\cdot)}^\bodyframe$ and ${(\cdot)}^\worldframe$ to indicate body-frame and world-frame, respectively.
Further, $\bar x$ defines a measurement and $\hat x$ defines an estimate or computed value of a quantity $x$. 
The rotation matrix from world-frame to body-frame is denoted as $\RotWtoB(\eulerAngles)\in\mathbf{R}^{3\times 3}$ as a function of the Euler angles~$\eulerAngles\in\mathbf{R}^3$.
Let $\0$ denote a zero matrix of zero vector of appropriate dimension.
Let $\I$ denote an identity matrix of appropriate dimension.

\subsection{Single-Model Kalman Filter}
\label{ssec:single_kalman}
A Kalman filter is a recursive state estimator 
for a dynamical system of the form:
\begin{subequations}
\begin{align} 
    \boldsymbol{x}_{t+1}
    &=
    \boldsymbol{A}_{t}
    \boldsymbol{x}_{t}
    +
    \boldsymbol{b}_{t} + \boldsymbol{w}_t 
    \\
    \boldsymbol{y}_{t} &= \boldsymbol{C}_{t}\boldsymbol{x}_{t}
    +
    \boldsymbol{v}_{t},
\end{align} 
\end{subequations}
where $\boldsymbol{x}_t\in\mathbf{R}^{n_x}$ is the state at discrete-time~$t$, $\boldsymbol{b}_t$ is the input, $\boldsymbol{y}_t\in\mathbf{R}^{n_y}$ is the measurement vector, $ \boldsymbol{w}_t \sim\mathcal{N}(\0,\boldsymbol{Q}_t)$ defines process noise, and $ \boldsymbol{v}_t \sim\mathcal{N}(\0,\boldsymbol{R}_t)$ defines measurement noise. 
At each recursion, the Kalman filter executes a prediction step and an update step~\cite{welch1995introduction}.
\begin{subequations}
\subsubsection{Prediction step}
In the prediction step, a model of the dynamical systems' evolution is used to propagate/predict the first and the second moment of the posterior distribution:
\begin{align}
        \hat{\boldsymbol{x}}_{t|t-1}
        &=
        {\boldsymbol{A}}_{t-1}
        \hat{\boldsymbol{x}}_{t-1|t-1}
        +
        {\boldsymbol{b}}_{t-1}
        \\
        {\boldsymbol{P}}_{t|t-1}
        &=
        {\boldsymbol{A}}_{t-1}
        {\boldsymbol{P}}_{t-1|t-1}
        {\boldsymbol{A}}_{t-1}^T
        +
        \boldsymbol{Q}_{t-1},
\end{align}
where $\hat{\boldsymbol{x}}_{t|t}$ and ${\boldsymbol{P}}_{t|t}$ are the mean and covariance of the distribution describing the state estimate at time~$t$, and $\hat{\boldsymbol{x}}_{t|t-1}$ and ${\boldsymbol{P}}_{t|t-1}$ are the predicted mean and covariance.

\subsubsection{Update step}
In the update step, the predicted mean and covariance are updated using sensor measurements:
    \begin{align}
        \tilde{\boldsymbol{y}}_t
        &=
        {\boldsymbol{y}}_t
        -
        {\boldsymbol{C}}_t
        \hat{\boldsymbol{x}}_{t|t-1}
        \\
        \boldsymbol{S}_t
        &=
        \boldsymbol{C}_t
        \boldsymbol{P}_{t|t-1}
        \boldsymbol{C}_t^T
        +
        \boldsymbol{R}_t
        \\
        \boldsymbol{K}_t
        &=
        \boldsymbol{P}_{t|t-1}
        \boldsymbol{C}_t^T
        \boldsymbol{S}_t^{-1}
        \\
        \hat{\boldsymbol{x}}_{t|t}
        &=
        \hat{\boldsymbol{x}}_{t|t-1}
        +
        \boldsymbol{K}_t
        \tilde{\boldsymbol{y}}_t
        \\
        \boldsymbol{P}_{t|t}
        &=
        \left(
        \I - 
        \boldsymbol{K}_t
        \boldsymbol{C}_t
        \right) 
        \boldsymbol{P}_{t|t-1}
    \end{align}
    with the innovation residual $\tilde{\boldsymbol{y}}_t$, the innovation covariance~$\boldsymbol{S}_t$, and the Kalman gain~$\boldsymbol{K}_t$.
\end{subequations}

\subsection{Interacting Multiple-Model Kalman Filter}
\label{ssec:immkf}
\begin{subequations}
An IMM-KF is a state estimator for a switched system of the form:
\begin{align}
    \boldsymbol{x}_{t+1}
    &=
    \boldsymbol A_t^{(k)} \boldsymbol{x}_t
    + 
    \boldsymbol{b}_t^{(k)} + \boldsymbol{w}_t^{(k)}
    \\
    \boldsymbol{y}_t
    &=
    \boldsymbol{C}_t^{(k)}\boldsymbol{x}_t
    + \boldsymbol{v}_t^{(k)},
\end{align}
where $k\in\{1,...,M\}$ denotes the $k^{\rm th}$ of the $M$ modes.
The switched system's modes are modeled as a Markov chain with with transition probabilities
\begin{align}
    \prob
    \left(m_{t+1}=m^{(j)} \middle|
    m_{t}=m^{(i)}\right)
    =
    \pi_{ij},
\end{align}
where
$m_{t}\in\mathcal{M}$ is the mode at time $t$, and the transition from mode~$m^{(i)}$ to mode~$m^{(j)}$ being~$\pi_{ij}$.
The IMM-KF involves an interaction, a filtering, a probability update, and a combination step~\cite{lefkopoulos2020interaction}.
\end{subequations}


\newcommand{\raisedMode}{{(k)}}
\subsubsection{Interaction step} 
\begin{subequations}
In the interaction step, the estimates of the $M$ filters are mixed and used to initialize each filter:
\begin{align}
    c^{(k)}
    &=
    \sum_{j=1}^{M}
    \pi_{jk} \mu_{t-1}^{(j)}
    \\
    \mu_{t-1|t-1}^{(j|k)}
    &=
    \frac{\pi_{jk}\mu_{t-1}^{(j)}}{c^{(k)}}
    \\
    \bar{\boldsymbol x}_{t-1|t-1}^{(k)}
    &=
    \sum_{j=1}^{M} 
    \mu_{t-1|t-1}^{(j|k)}
    \hat{\boldsymbol x}_{t-1|t-1}^{(j)}
    \\
    \bar{\boldsymbol{P}}_{t-1|t-1}^{(k)}
    &=
    \sum_{j=1}^{M} 
    \mu_{t-1|t-1}^{(j|k)}
    \left(
    \boldsymbol{P}_{t-1|t-1}^{(j)}
    +
    \boldsymbol{X}_{t-1|t-1}^{(k,j)}
    \right)
\end{align}
with $\boldsymbol{X}_{t|t}^{(k,j)}= \big( \bar{\boldsymbol{x}}_{t|t}^{(k)} - \hat{\boldsymbol{x}}_{t|t}^{(j)} \big)
\big( \bar{\boldsymbol{x}}_{t|t}^{(k)} - \hat{\boldsymbol{x}}_{t|t}^{(j)} \big)^T$,
where $\hat{\boldsymbol x}_{t|t}^{(k)}$ is the state estimate of filter~$k$ at time~$t$,
and
$\mu_{t}^{(k)}$ is the probability of filter~$k$ being active.

\subsubsection{Filtering step} 
Each Kalman as in Section~\ref{ssec:single_kalman} is executed separately using  the mixed estimate $\bar{\boldsymbol x}_{t-1|t-1}^{(k)}$ and $\bar{\boldsymbol{P}}_{t-1|t-1}^{(k)}$ to initialize filter~$k$.

\subsubsection{Probability update step} 
The filters' innovation residuals are used to update the filters' model probabilities: 
\begin{align}
\label{eq:mode_likelihoods}
    L_t^\raisedMode
    &=
    \frac{
    \exp{\left(
    -\frac{1}{2} 
     \big(\tilde{\boldsymbol y}_t^\raisedMode\big)^T 
    \big(\boldsymbol S_t^\raisedMode \big)^{-1} 
    \tilde{\boldsymbol y}_t^\raisedMode
    \right)
    }}{
    \left|2\pi \boldsymbol S_t^{(k)}\right|^{0.5}
    }
    \\
    \mu_t^\raisedMode
    &=
    \frac{c^\raisedMode L_t^\raisedMode}{
    \sum_{i=1}^M c^{(i)} L_t^{(i)}
    }
\end{align}
with the likelihood $L_t^\raisedMode$ and the model probability $\mu_t^\raisedMode$.



\subsubsection{Combination step} 
Lastly, the filters' state estimates are combined as a weighted sum using the model probabilities:
\begin{align}
    \hat{\boldsymbol{x}}_{t|t}
    &=
    \sum_{k=1}^M
    \mu_t^\raisedMode \hat{\boldsymbol{x}}_{t|t}^\raisedMode
    \\
    {\boldsymbol{P}}_{t|t}
    &=
    \sum_{k=1}^M
    \mu_t^\raisedMode 
    \left(
    {\boldsymbol{P}}_{t|t}^\raisedMode
    +
    \big(
     \hat{\boldsymbol{x}}_{t|t}
     - 
     \hat{\boldsymbol{x}}_{t|t}^\raisedMode
    \big)
    \big(
     \hat{\boldsymbol{x}}_{t|t}
     - 
     \hat{\boldsymbol{x}}_{t|t}^\raisedMode
    \big)^T
    \right).
\end{align}
\end{subequations}

\section{Mode-dependent Motion Model} 
\label{sec:math_derive}
The motion of the robot's trunk can be modeled in continuous time as 
\begin{subequations}
\begin{align}
\label{eq:baseline_cont_dyn}
    \dot{\boldsymbol{x}} (t) 
    = 
    \boldsymbol{A}(\eulerAngles(t))
    \boldsymbol{x}(t)
    +
    \boldsymbol{g}  
    +
    \boldsymbol{b} (t)
\end{align} 
with the state $\boldsymbol{x}(t)\in\mathbf{R}^{12}$ at time $t$, the gravity vector $\boldsymbol{g}$, 
and $\boldsymbol{b}(t)$ resulting from external forces.
The state consists of the orientation $\eulerAngles$, the center of mass (CoM) position $\boldsymbol{d}_{\rm CoM}^\worldframe$, 
the rotation $\rotationsWF$,
and the linear velocity $\boldsymbol{v}_{\rm CoM}^\worldframe$:
\begin{align}
    \boldsymbol{x}(t)
    =
    \begin{bmatrix}
    \eulerAngles(t)
    \\
    \boldsymbol{d}_{\rm CoM}^\worldframe(t)
    \\
    \rotationsWF(t)
    \\
    \boldsymbol{v}_{\rm CoM}^\worldframe(t)
    \end{bmatrix}.
\end{align}
Hence,
\begin{align}
    \boldsymbol{A} (\eulerAngles(t))
    =
    \begin{bmatrix}
        \0 & \0 & \RotWtoB(\eulerAngles(t)) & \0
        \\
        \0 & \0 & \0 & \I 
        \\
        \0 & \0 & \0 & \0 
        \\
        \0 & \0 & \0 & \0 
    \end{bmatrix}.
\end{align}
\end{subequations}

The external forces result from contacts of the robot's feet with the ground/environment, i.e.,
\begin{subequations}
    \begin{align}
        \boldsymbol{b}(t)
        = 
        \boldsymbol{B}(t)
        \boldsymbol{f}^{\bodyframe}(t)
    \end{align}
    with the force vector $\boldsymbol{f}^{\bodyframe} (t)$. 
    For a four-legged robot,
    \begin{align} 
    \boldsymbol{f}^{\bodyframe} (t)= 
    \begin{bmatrix}
        \boldsymbol{f}_{\RF}^\bodyframe (t)
        \\
        \boldsymbol{f}_{\LF}^\bodyframe (t)
        \\
        \boldsymbol{f}_{\RR}^\bodyframe (t)
        \\
        \boldsymbol{f}_{\LR}^\bodyframe(t)
    \end{bmatrix},
    \end{align}
    where $\boldsymbol{f}_i^\bodyframe = [{f}_{x,i}^\bodyframe\quad {f}_{y,i}^\bodyframe\quad {f}_{z,i}^\bodyframe]^T \in\mathbf{R}^3$ is the foot-contact force of leg~$i$. The reason for representing the contact forces in body-frame will become apparent in Section~\ref{sec:estimation}.
    Hence,  
    \begin{align*}
    \boldsymbol{B}(t)
    &=
    \boldsymbol{B}_1(t)\boldsymbol{B}_2(t)
    \\
        \boldsymbol{B}_1(t)
        &=
        \begin{bmatrix}
            \0 & \0
            \\
            \0 & \0
            \\
            \RotBtoW (\eulerAngles(t)) \left(\inertiaBF\right)^{-1} & \0
            \\
            \0 & \RotBtoW (\eulerAngles(t)) \frac{1}{m}
        \end{bmatrix} 
        \\
    \boldsymbol{B}_2(t)
    &=
    \begin{bmatrix}
        \boldsymbol{P}_{\RF}^\bodyframe(t)
        &
        \boldsymbol{P}_{\LF}^\bodyframe(t)
        &
        \boldsymbol{P}_{\RR}^\bodyframe(t)
        &
        \boldsymbol{P}_{\LR}^\bodyframe(t)
        \\
        \I & \I & \I & \I
        \end{bmatrix},
    \end{align*}
    where $\boldsymbol{B}_1(t)$ includes rotations and inertia values of the robot's trunk, 
    and $\boldsymbol{B}_2(t)$ computes linear and angular accelerations from the contact forces with 
    \begin{align}
        \boldsymbol{P}_{i}^\bodyframe(t) = 
        {\boldsymbol{P}_{\rm skew}}(\boldsymbol{p}_{i}^\bodyframe(t) 
        ) \in\mathbf{R}^{3\times 3},
    \end{align}
    where $\boldsymbol{p}_{i}^\bodyframe \in\mathbf{R}^{3}$ is the  relative position of leg~$i$ from the trunk's CoM, which can be calculated using forward kinematics and the joint angles $\boldsymbol{\legJointPosSymbol}_i$, i.e., 
    $\boldsymbol{p}_{i}^\bodyframe(t)= \FK_i (\boldsymbol{\legJointPosSymbol}_i)$.
    Hence, $\boldsymbol{B}_2(t)\boldsymbol{f}^{\bodyframe} (t)$ computes the resulting moments and forces at the CoM (in body-frame) as a combination of the forces at the feet, which are transformed to angular and linear accelerations (in world-frame) via $\boldsymbol{B}_1(t)$.
\end{subequations}

\subsection{Model as Switched System} 
As the contact forces $\boldsymbol{f}_{i}^\bodyframe$ are often difficult to measure, this paper proposes to use estimated ``hypothetical" forces in combination with binary contact-detection variables, $\delta_i^k$, with mode index~$k$:
\begin{align*}
    \boldsymbol{f}_{i}^\bodyframe(t)
    =
    \delta_i^k
    \est{\boldsymbol{f}}_{i}^\bodyframe (t).
\end{align*}
The estimated ``hypothetical" forces, $\hat{\boldsymbol{f}}_{i}^\bodyframe$, indicate a contact force acting on the robot if the robot's leg {\it{would be}} in contact with the ground. 
Hence, if the leg~$i$ is actually in contact with ground, the binary variable has to be one, $\delta_i^k=1$, for the dynamics to be accurate. In turn, this observation can be leveraged to detect feet that are in contact with the ground.
Hence, simultaneous contact detection and state estimation is achieved by jointly estimating $\boldsymbol{x}(t)$ and $\delta^k_i$.
Thus, the dynamics in~\eqref{eq:baseline_cont_dyn} are represented by a switched system with
\begin{align}
\label{eq:cont_time}
    \dot{\boldsymbol{x}}(t)
    =
    \boldsymbol{A}(\eulerAngles(t)) \boldsymbol{x}(t)
    +
    \boldsymbol{g}  
    +
    \boldsymbol{B}^k(t)
    \est{\boldsymbol{f}}^\bodyframe (t),
\end{align}
where the computation/estimation of $\est{\boldsymbol{f}}^\bodyframe$ is discussed in Section~\ref{sec:estimation},
and
\begin{align*}  
\boldsymbol{B}^k(t) &= \boldsymbol{B}_1(t)\boldsymbol{B}_2^k(t)
    \\
    \boldsymbol{B}_2^k
    &=
    \begin{bmatrix}
        \delta_{\RF}^k \boldsymbol{P}_{\RF}^\bodyframe
        &
        \delta_{\LF}^k\boldsymbol{P}_{\LF}^\bodyframe
        &
        \delta_{\RR}^k\boldsymbol{P}_{\RR}^\bodyframe
        &
        \delta_{\LR}^k\boldsymbol{P}_{\LR}^\bodyframe
        \\
        \delta_{\RF}^k \I & \delta_{\LF}^k \I & \delta_{\RR}^k \I & \delta_{\LR}^k \I
        \end{bmatrix}.
\end{align*}

\subsection{Contact Modes and Contact Probabilities}
A complete enumeration of modes for a four-legged robot is:
\begin{align*}
    k &=1, m^{(1)} \rightarrow
    \delta_{\RF}^1 = 0,
    \delta_{\LF}^1 = 0,
    \delta_{\RR}^1 = 0,
    \delta_{\LR}^1 = 0
    \\
    k &=2, m^{(2)} \rightarrow
    \delta_{\RF}^2 = 0,
    \delta_{\LF}^2 = 0,
    \delta_{\RR}^2 = 0,
    \delta_{\LR}^2 = 1
    \\
    k &=3, m^{(3)} \rightarrow
    \delta_{\RF}^3 = 0,
    \delta_{\LF}^3 = 0,
    \delta_{\RR}^3 = 1,
    \delta_{\LR}^3 = 0
    \\
    &\vdots
    \\
    k &=15, m^{(15)} \rightarrow
    \delta_{\RF}^{15} = 1,
    \delta_{\LF}^{15} = 1,
    \delta_{\RR}^{15} = 1,
    \delta_{\LR}^{15} = 0
    \\
    k &=16, m^{(16)} \rightarrow
    \delta_{\RF}^{16} = 1,
    \delta_{\LF}^{16} = 1,
    \delta_{\RR}^{16} = 1,
    \delta_{\LR}^{16} = 1
\end{align*}
with the total number of modes $M$.
For example, $m^{(1)}$ and $m^{(16)}$ indicates none and all four legs are in contact with the ground, respectively.
Note that for computational reasons, some less likely modes can be neglected, e.g., $k\in\{2,3,5,9\}$ relate to only one leg being in contact with the ground, which is less likely during standard operation.

Using a probabilistic state estimator with the switched system in~\eqref{eq:cont_time}, it is further possible to indicate probabilities of leg~$i$ being in contact with the ground.
Let $\mu^k$ be the likelihood that mode $m^{(k)}$ is active, see the preliminaries in Section~\ref{ssec:immkf}. 
Then, the likelihood of leg~$i$ being in contact with the ground can be calculated as
\begin{align}
\label{eq:calc_contact_prob}
    p_i = \prob(\delta_i=1) = \sum_{k=1}^M \mu^k \delta_i^k.
\end{align}
The algorithm in this paper leverages these probabilities in combination with joint angles and joint velocities to estimate the trunk's CoM position and velocity, which is discussed in the following section.

\section{Estimation and Filter Design}
\label{sec:estimation}
For state estimation, this paper assumes the availability of the following sensor measurements
\begin{itemize}
    \item Euler angles~$\measEulerAngles$, e.g., provided by the IMU
    \item Acceleration at CoM~$\measAccelerationsBF$, e.g., provided by the IMU
    \item Rotations at CoM~$\measRotationsBF$, e.g., provided by the IMU
    \item Joint angles~$\measLegJointPos$, e.g., provided by motors
    \item Joint angle velocities~$\measLegJointVel$, e.g., provided by motors
    \item Joint torques~${\boldsymbol{\tau}}$, e.g., provided by motors
\end{itemize}
Further, this paper uses first-order Euler discretization with sampling rate $T_s$ of the switched system in~\eqref{eq:cont_time},
\begin{align}
    \boldsymbol{x}_{t+1}
    =
    \boldsymbol{x}_{t}
    +
    T_s
    \left(
    \boldsymbol{A}(\eulerAngles_t) \boldsymbol{x}_t
    +
    \boldsymbol{g}  
    +
    \boldsymbol{B}^k_t
    \est{\boldsymbol{f}}^\bodyframe_t
    \right)
    +\boldsymbol{w}_t
\end{align}
with the process noise $\boldsymbol{w}_t\sim\mathcal{N}(\0,\boldsymbol{Q}_t)$,
where sub-script $t$ is used to indicate discrete time.

\begin{remark}
    The joint angle velocities can also be estimated if they are not directly measured.
\end{remark}

\subsection{Pseudo Measurements}
Without a positioning system, the robot does not provide a measurement of the trunk's position or the velocity.
However, if the feet being in contact with the ground are known, then kineamtic relationships can be leveraged to provide a ``pseudo measurement" for the trunk's position and velocity.
Hence, detecting contacts while estimating the state enables the use of such pseudo measurements.

In order to obtain pseudo measurements for the robot's trunk velocities, the mode-probabilities in~\eqref{eq:calc_contact_prob} can be leveraged.
Let ${\boldsymbol{v}}^{{\bodyframe}}_i = \boldsymbol{J}_i(\boldsymbol{\legJointPosSymbol}_i)\dot{\boldsymbol{\legJointPosSymbol}}_i$ be the foot-velocity of leg~$i$ in reference to the robot's CoM (in body-frame). 
Then, if leg~$i$ is in contact with the ground (and not slipping),
${\boldsymbol{v}}^{{\worldframe}}_i = \0$, which implies ${\boldsymbol{v}}^{{\worldframe}}_{\rm CoM} = - \RotBtoW {\boldsymbol{v}}^{{\bodyframe}}_i$.
Hence, pseudo measurements can be obtained by considering contacts and footstep velocities.
In this paper, the probabilities in~\eqref{eq:calc_contact_prob} are used to obtain such pseudo measurements as
\begin{subequations}
\begin{align}
    \pseudoMeas{\boldsymbol{v}}^{\worldframe}_{\rm CoM} = 
    -
    \frac{1}{
    \sum_{i=1}^4 
    h(p_i)
    }
    \sum_{i=1}^4
    h(p_i)
    \RotBtoW(\estEulerAngles)
    \pseudoMeas{\boldsymbol{v}}^{\bodyframe}_i
\end{align}
with $\pseudoMeas{\boldsymbol{v}}^{{\bodyframe}}_i = \boldsymbol{J}_i(\estLegJointPos_i)\estLegJointVel_i$,
where $h$ is a function that computes a weight based on the contact probability~$p_i$.
This paper uses $h(p_i)=\max(p_i-\bar{p},0)$ with the threshold value $\bar{p}=60\%$.
The reason for choosing such a threshold value is the exclusion of legs that are likely not in contact with the ground for computing the pseudo measurements.

Similarly, for obtaining a position measurement,
\begin{align}
\label{eq:pseudo_pos}
    \pseudoMeas{\boldsymbol{p}}^{\worldframe}_{\rm CoM} = 
    -
    \frac{1}{
    \sum_{i=1}^4 
    h(p_i)
    }
    \sum_{i=1}^4
    h(p_i)
    \RotBtoW(\estEulerAngles)
    \pseudoMeas{\boldsymbol{p}}^{\bodyframe}_i
\end{align}
with $\pseudoMeas{\boldsymbol{p}}^{{\bodyframe}}_i = \FK_i (\estLegJointPos_i)$.
\end{subequations}


\begin{remark}
    The advantage of using~\eqref{eq:pseudo_pos} is that the robot's vertical position, $p_{z,{\rm CoM}}^{\worldframe}$, is estimated in reference to the ground rather than in global coordinates, which is useful for control algorithms. 
    The horizontal position defined by $p_{x,{\rm CoM}}^{\worldframe}$ and $p_{y,{\rm CoM}}^{\worldframe}$ is estimated in reference to the footstep positions, i.e., if $p_{x,{\rm CoM}}^{\worldframe}=p_{y,{\rm CoM}}^{\worldframe}=0$, then the robot's CoM is perfectly in the geometric center of the feet, which is also useful for control.
\end{remark}

\subsection{Hypothetical Forces}

For the four-legged robot configuration considered in this paper, the hypothetical forces $\hat{\boldsymbol{f}}_{i}^\bodyframe$ can be approximated using using the foot Jacobian $\Jac_i$ with the joint angles $\boldsymbol{q}_i$ and the motor torques $\boldsymbol{\tau}_i$ of leg~$i$: 
\begin{align}
\label{eq:compute_forces}
    \est{\boldsymbol{f}}_{i,t}^\bodyframe
    = 
    (\Jac_i (\boldsymbol{q}_{i,t}))^{-1}
    \boldsymbol{\tau}_{i,t},
\end{align}
which is a common practice for legged robots, see, e.g.,~\cite{di2018dynamic, schperberg2022auto} where this relationship is used for a stance controller.
Note that during standard operation, $\Jac_i$ is invertible for a leg with three joints/motors.

\subsection{Measurement Equation}
The estimation algorithm in this paper uses the measurements $\boldsymbol{y}_t\in\mathbf{R}^{15}$ with
\begin{align}
\boldsymbol{y}_t
    =
    \boldsymbol{C}(\eulerAngles_t)
    \boldsymbol{x}_t
    +
    \boldsymbol{D}^k
    \est{\boldsymbol{f}}^\bodyframe_t
    +
    \boldsymbol{v}_t
\end{align}
with the sensor noise $\boldsymbol{v}_t\sim\mathcal{N}(\0,\boldsymbol{R}_t)$,
\begin{align*}
    \boldsymbol{y}_t
    &=
    \begin{bmatrix}
        \meas{\eulerAngles}_t
        \\
        \pseudoMeas{\boldsymbol{p}}^\worldframe_t
        \\
        \measRotationsBF_t
        \\
        \pseudoMeas{\boldsymbol{v}}^\worldframe_t
        \\
        \meas{\boldsymbol{a}}^\bodyframe_t
    \end{bmatrix},\,
    \boldsymbol{C}(\eulerAngles_t)
    =
    \begin{bmatrix}
        \I  & \0 & \0 & \0
        \\
        \0 & \I & \0 & \0
        \\
        \0 & \0 & \RotWtoB (\eulerAngles_t) & \0
        \\
        \0 & \0 & \0 & \I
        \\
        \0 & \0 & \0 & \0
    \end{bmatrix}
\end{align*}
and  
\begin{align*}
    \boldsymbol{D}^k
    =
    \begin{bmatrix}
        \0  & \0 & \0 & \0
        \\
        \0 & \0 & \0 & \0
        \\
        \0 & \0 & \0 & \0
        \\
        \0 & \0 & \0 & \0
        \\
        \delta^k_{\RF} \frac{1}{m}\I & \delta^k_{\LF} \frac{1}{m}\I & \delta^k_{\RR} \frac{1}{m}\I  & \delta^k_{\LR} \frac{1}{m}\I 
    \end{bmatrix}
\end{align*}
The reason for $\boldsymbol{D}^k$ being mode dependent is that measured accelerations at the CoM are the sum of the contact forces with $\delta^k_{i}=1$, i.e., of feet in contact with the ground.
While the acceleration measurements in this setting do not directly impact the state estimate, they help in determining the ``correct" mode of operation.
Hence, an advantage of the proposed algorithm is that it can leverage acceleration measurements, $\meas{\boldsymbol{a}}^\bodyframe_t$, for state estimation by means of altering the modes' probabilities. 

\subsection{Overall Filter Design}
Overall, the process model and measurement model for each Kalman filter related to mode $m^{(k)}$ are given by
\begin{subequations}
    \begin{align}
    \label{eq:switched_sys}
        \hat{\boldsymbol{x}}_{t+1}
         & =
        \left(
        \I_{12}
        +
        T_s 
        \boldsymbol{A}(\estEulerAngles_t) \right) \hat{\boldsymbol{x}}_{t}
        +
        \boldsymbol{g}  
        +
        \boldsymbol{B}^k_t
        \est{\boldsymbol{f}}^\bodyframe_t 
        +
        \boldsymbol{w}_t
        \\
        \boldsymbol{y}_t &= \boldsymbol{C}(\estEulerAngles_t) \hat{\boldsymbol{x}}_{t}
        +
        \boldsymbol{D}^k
        \est{\boldsymbol{f}}^\bodyframe_t
        +
        \boldsymbol{v}_t.
    \end{align}
\end{subequations}

\begin{remark}[Confidence in Pseudo Measurements]
The algorithm leverages time-varying covariance matrices $\boldsymbol{Q}_t, \boldsymbol{R}_t$ in order to adjust the confidence in the pseudo-measurements.
This paper scales the elements in $\boldsymbol{R}_t$ that relate to the pseudo measurements with
\begin{align}
\label{eq:pseudo_conf}
    r_{\rm scale} = \frac{1}{1 + 100\sum_{i=1}^4 h(p_i)},
\end{align}
i.e., for higher confidence in a foot being in contact with the ground, the algorithm decreases the uncertainty associated with the measurement.
Hence, the contact probabilities in~\eqref{eq:calc_contact_prob} are useful not only to compute a pseudo measurement but also to reason about its confidence.
\end{remark}


\subsection{Exploiting Physical Limits of Contact Forces for Biasing Mode Probabilities}
\label{ssec:force_positive}

There are two physical limits that can be exploited easily for more accurate contact detection and state estimation.
First, contact forces need to be positive, because feet can push the robot up but cannot pull the robot down.
Hence, if $\hat f_{z,i}^{\worldframe}<0$, then all modes~$k$ with $\delta_i^k=1$ are less likely.
Second, horizontal forces are constrained by friction to be smaller (in absolute value) than $\nu \hat f_{z,i}^{\worldframe}$ with the friction coefficient $0\leq \nu \leq 1$, i.e., $((\hat f_{x,i}^{\worldframe})^2+(\hat f_{y,i}^{\worldframe})^2)^{0.5}< \nu \hat f_{z,i}^{\worldframe}$.

An IMM-KF allows to incorporate such prior knowledge/physical limits by means of biasing the modes' probabilities. 
Instead of~\eqref{eq:mode_likelihoods}, this paper uses
\begin{align*}
    &L_t^\raisedMode
    =
    \frac{
    \exp{\left(
    -\frac{1}{2} 
     \big(\tilde{\boldsymbol y}_t^\raisedMode\big)^T 
    \big(\boldsymbol S_t^\raisedMode \big)^{-1} 
    \tilde{\boldsymbol y}_t^\raisedMode
    \right)
    }}{
    \left|2\pi \boldsymbol S_t^{(k)}\right|^{0.5}
    }
    \exp{
    \left(
    -
    h_f\left(\est{\boldsymbol{f}}^\worldframe_t\right)
    \right)}
    \\
    &
    h_f(\est{\boldsymbol{f}}^\bodyframe_t)
    =
    c_{\rm force}
    \sum_{i=1}^4
    \delta^k_i
    \left(\min 
    \left(
    0, \hat f_{z,i}^{\worldframe}
    \right)\right)^2
\end{align*}
with some weight $c_{\rm force}$.

\begin{remark}[Alternative Formulation] 
    An alternative to using forces as inputs to the dynamical system in~\eqref{eq:switched_sys} is to augment the state by contact forces.
    This may lead to more accurate estimation of contact forces, but comes at the expense of higher computation cost. 
    Such a formulation with ``forces as pseudo states" may be useful in cases, where~\eqref{eq:compute_forces} does not provide a good estimate for the forces. 
\end{remark}

\section{Results} 
\label{sec:results}
\subsection{Experimental Setup} 
\label{sec:setup}
\begin{figure}[t]
    \centering
    \includegraphics[width=.95\columnwidth]{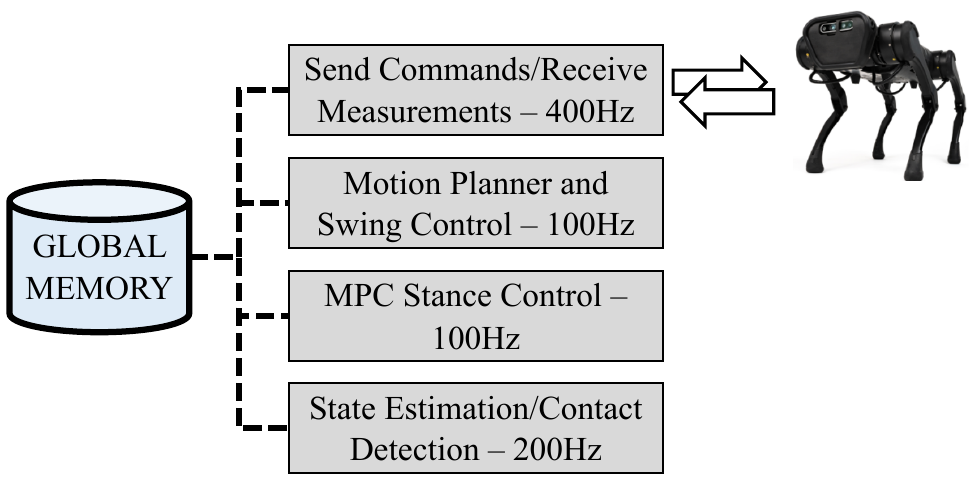} 
    \caption{
    Controller architecture.
    For the Gazebo validations, the controller publishes ROS topics that the robot uses.
    For the hardware experiments, the controller communicates with the robot via Ethernet. 
    }
    \label{fig:controller}
\end{figure}

Fig.~\ref{fig:controller} summarizes the controller architecture and communication with the robot for both Gazebo and hardware experiments.
The controller executes four modules in parallel, whose execution rate are displayed in Fig.~\ref{fig:controller}.
The four modules share a global memory, which they use to communicate their respective results.
First, one module links the controller with the robot. 
This module sends motor commands from the global memory to the robot and receives measurements from the robot, which are subsequently written into the memory.
The second module is used for motion planning and the coordination of the robot's legs.
The third module uses the state estimate stored in the global memory to compute motor commands that stabilize the robot.
The state estimation and contact detection module pulls the robot's measurements from the global memory, processes the measurements using the algorithm in this paper, and pushes the updated state estimate to the global memory.
This paper focuses primarily on the state estimation and contact detection module.
Note that the proposed algorithm can be utilized for (almost) any controller structure.
For both Gazebo simulation and hardware experiments, torque commands are sent using the robotic operating system (ROS). 

\subsection{Design Choices}
For validation, the following modes are implemented:
\begin{align*}
    k &=1, m^{(1)} \rightarrow
    \delta_{\RF}^1 = 0,
    \delta_{\LF}^1 = 0,
    \delta_{\RR}^1 = 0,
    \delta_{\LR}^1 = 0   
    \\
    k &=2, m^{(2)} \rightarrow
    \delta_{\RF}^2 = 0,
    \delta_{\LF}^2 = 1,
    \delta_{\RR}^2 = 1,
    \delta_{\LR}^2 = 0   
    \\
    k &=3, m^{(3)} \rightarrow
    \delta_{\RF}^3 = 0,
    \delta_{\LF}^3 = 1,
    \delta_{\RR}^3 = 1,
    \delta_{\LR}^3 = 1   
    \\
    k &=4, m^{(4)} \rightarrow
    \delta_{\RF}^4 = 1,
    \delta_{\LF}^4 = 0,
    \delta_{\RR}^4 = 0,
    \delta_{\LR}^4 = 1   
    \\
    k &=5, m^{(5)} \rightarrow
    \delta_{\RF}^5 = 1,
    \delta_{\LF}^5 = 0,
    \delta_{\RR}^5 = 1,
    \delta_{\LR}^5 = 1   
    \\
    k &=6, m^{(6)} \rightarrow
    \delta_{\RF}^6 = 1,
    \delta_{\LF}^6 = 1,
    \delta_{\RR}^6 = 0,
    \delta_{\LR}^6 = 1   
    \\
    k &=7, m^{(7)} \rightarrow
    \delta_{\RF}^7 = 1,
    \delta_{\LF}^7 = 1,
    \delta_{\RR}^7 = 1,
    \delta_{\LR}^7 = 0   
    \\ 
    k &=8, m^{(8)} \rightarrow
    \delta_{\RF}^{8} = 1,
    \delta_{\LF}^{8} = 1,
    \delta_{\RR}^{8} = 1,
    \delta_{\LR}^{8} = 1
\end{align*}
Here, the modes $k=2,4$ relate to a trot gait.
The modes $k=3,5,6,7$ relate to three feet being in contact with the ground, which are important for walking gait and for early contact detection during trot gait.

The transition probabilities of the IMM-KF are chosen for trot gait being the standard mode operation as
\newcommand{\pnone}{\pi_0}
\newcommand{\ptrot}{\pi_1}
\newcommand{\pstand}{\pi_3}
\newcommand{\pthree}{\pi_2}
\begin{align*}
    &\boldsymbol{\pi}=\\
    &\begin{bmatrix} 
        \pnone & \frac{1\!-\!\pnone}{2} & 0 & \frac{1\!-\!\pnone}{2} & 0 & 0 & 0 & 0
        \\
        \frac{1\!-\!\ptrot}{5} & \ptrot & \frac{1\!-\!\ptrot}{5} & \frac{1\!-\!\ptrot}{5} & 0 & 0 & \frac{1\!-\!\ptrot}{5} & \frac{1\!-\!\ptrot}{5}
        \\
        0 & \frac{1\!-\!\pthree}{2} & \pthree & 0 & 0 & 0 & 0 & \frac{1\!-\!\pthree}{2}
        \\
        \frac{1\!-\!\ptrot}{5} & \frac{1\!-\!\ptrot}{5} & 0 & \ptrot &  \frac{1\!-\!\ptrot}{5} & \frac{1\!-\!\ptrot}{5} & 0 & \frac{1\!-\!\ptrot}{5}
        \\
        0 & 0 & 0 & \frac{1\!-\!\pthree}{2} & \pthree & 0 & 0 & \frac{1\!-\!\pthree}{2}
        \\
        0 & 0 & 0 & \frac{1\!-\!\pthree}{2} & 0 & \pthree & 0 & \frac{1\!-\!\pthree}{2}
        \\
        0 & \frac{1\!-\!\pthree}{2} & 0 & 0 & 0 & 0 & \pthree & \frac{1\!-\!\pthree}{2}
        \\
        0 & \frac{1\!-\!\pstand}{2} & 0 & \frac{1\!-\!\pstand}{2} & 0 & 0 & 0 & \pstand
    \end{bmatrix}
\end{align*}
with $\pi_0=\pi_1=\pi_2=\pi_3=0.8$.
Note that these transition probabilities were designed for trot. However, walking gait, e.g., can be considered by changing the last row of the transition probability matrix. 
Finally, the IMM-KF uses $\boldsymbol{Q}=10^{-2}\cdot{\rm diag}([1,1,1,10,10,10,1,1,1,0.01,0.01,0.01])$ and $\boldsymbol{R}_t=10^{-4} \cdot {\rm diag}([1,1,1,10^{4}r_{\rm scale},10^{4}r_{\rm scale},10r_{\rm scale},1,1,1,$ $10^{4}r_{\rm scale},10^{4}r_{\rm scale},10^{4}r_{\rm scale},1,1,1])$ with the confidence of the pseudo measurements $r_{\rm scale}\leq 1$ as in~\eqref{eq:pseudo_conf}.

\subsection{Gazebo Simulation Results}
\begin{figure}[t]
    \centering
    \includegraphics[width=0.85\columnwidth]{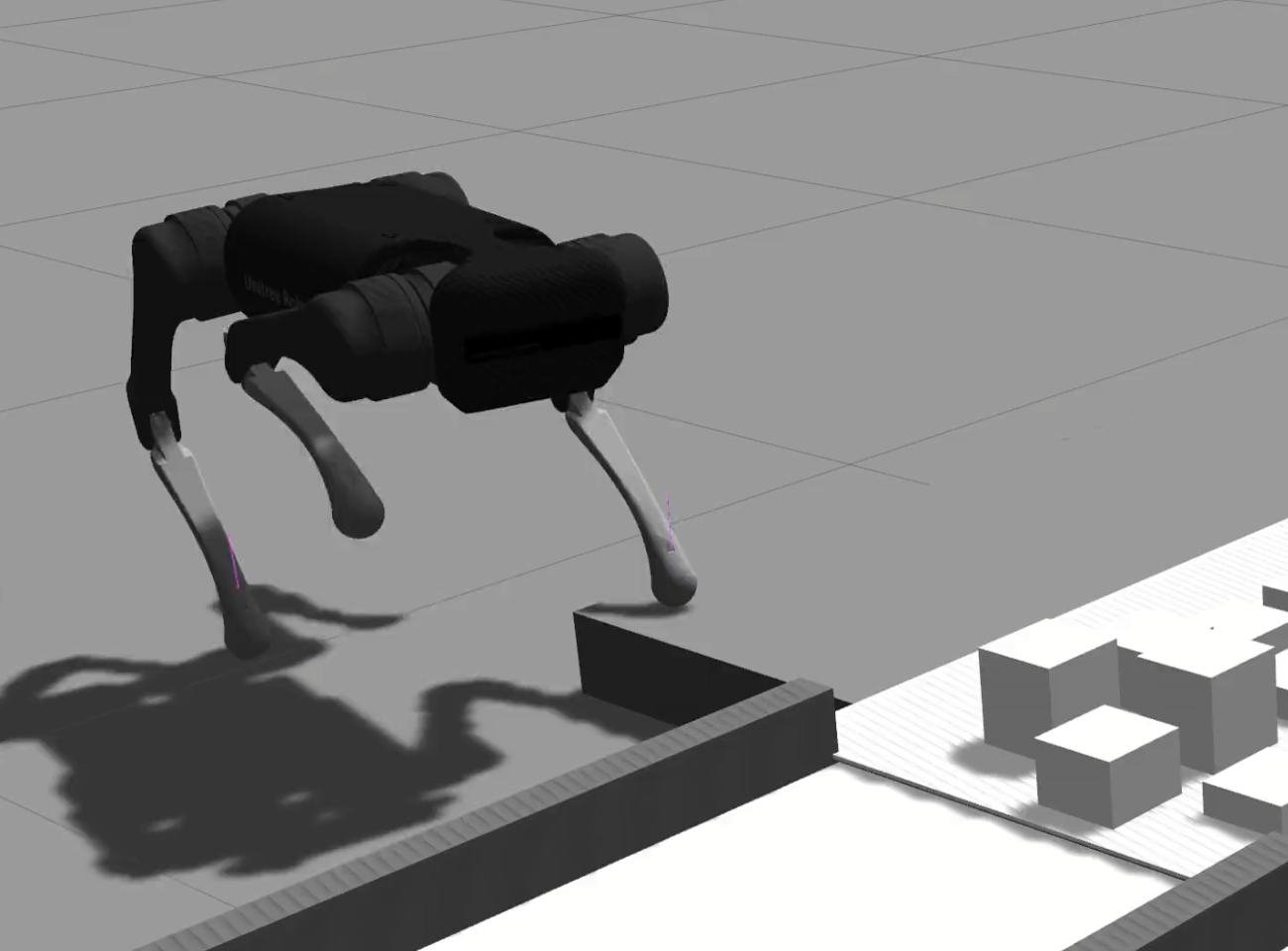} 
    \caption{
    Screenshot of Gazebo simulation environment. 
    }
    \label{fig:gazebo_screenshot}
\end{figure}

First, the algorithm is verified using a simulation study.
The rationale for including a simulation study is the presence of a ground truth, which is absent for hardware experiments.
The IMM-KF was implemented for the Unitree A1 robot~\cite{A1} using Gazebo with the bullet physics engine, see Fig.~\ref{fig:gazebo_screenshot} for an illustration.
The Unitree A1 simulator is a realistic high-fidelity environment, which is made available by the robot manufacturer. 
Further, the Gazebo simulation environment is executed in a parallel thread, and hence the software needs to be executed during run-time of the simulation. 
The proposed IMM-KF estimate is compared with a baseline Kalman filter, which uses the switched dynamics in~\eqref{eq:switched_sys} with the modes given by the controller instead of being estimated.

\begin{figure}[t]
    \centering
    \includegraphics[width=0.925\columnwidth]{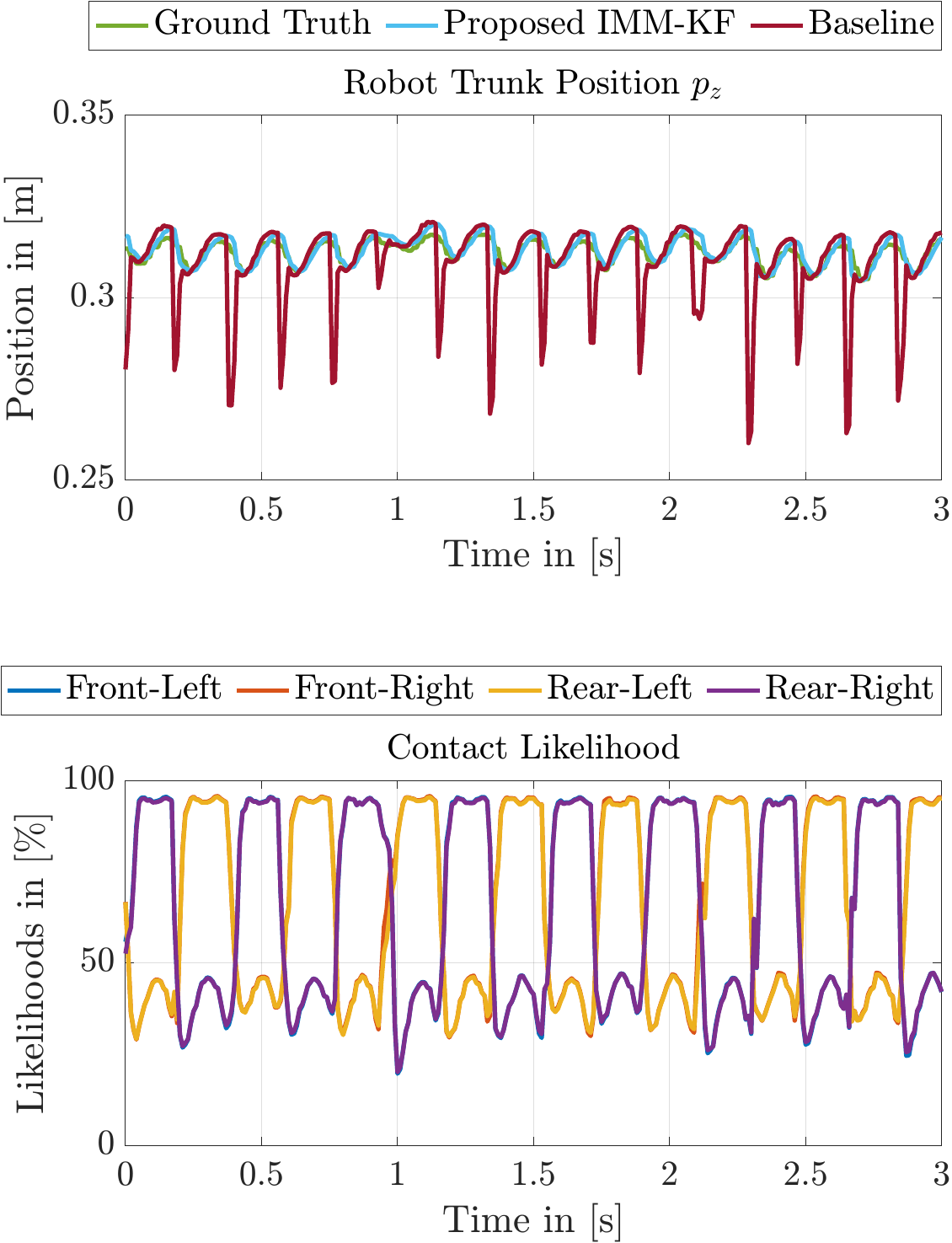} 
    \caption{Gazebo simulation results. 
    Top: Vertical position tracking. 
    It shows the ground truth (green), the estimate provided by the proposed IMM-KF (light blue), and the baseline (dark red).
    Bottom: Contact likelihoods of the four feet.
    }
    \label{fig:contact_gazebo}
\end{figure}

Fig.~\ref{fig:contact_gazebo} shows Gazebo simulation results for a 3s window.
The robot is traversing with a constant forward velocity of 1m/s in trot gait. 
As can be seen, the contact likelihoods of the four feet calculated as in~\eqref{eq:calc_contact_prob} indicate correctly that the front-left and the rear-right leg are swinging together, switching off with the front-right and rear-left.
Further, the algorithm is able to quickly identify switching contact, which is highlighted in more detail in Section~\ref{ssec:hardware}.
Fig.~\ref{fig:contact_gazebo} also shows that the proposed IMM-KF solution tracks the robot's vertical position closely, whereas the baseline spikes in between gait transitions.
These spikes are due to the baseline filter not actively identifying modes, but using the controller modes in open loop.
Table~\ref{tab:rmse} quantifies the tracking error for 1min of operation for the full state, the vertical position of the robot's CoM, and the robot's velocities.
As displayed in Fig.~\ref{fig:contact_gazebo}, the robot's vertical position above ground is crucial to stabilize the robot, especially for controllers that use torque commands.
\begin{table}[t]
    \centering
    \caption{
    State Tracking: Root Mean Square Error
    }
    \begin{tabular}{cc|c|c|c}  
     & & IMM-KF & Baseline \\
        \toprule 
        Full State & RMSE
         & 0.0952 & 0.2438  \\ 
         \midrule
        \multirow{2}{*}{Vertical Position} & RMSE
         & 0.17cm & 1.25cm \\
        & Maximum & 0.88cm & 6.36cm \\ 
         \midrule
         Velocities & RMSE
         & 0.1195m/s & 0.4395m/s \\  
        \bottomrule
    \end{tabular}
    \label{tab:rmse}
\end{table} 
Table~\ref{tab:rmse} shows that the state tracking error is reduced by a factor of 2.5.
For the vertical position, the tracking error is reduced by 7.5, and the velocity tracking error is reduced by 3.75.
These improvement highlight the benefit of simultaneously estimating the state and detecing contacts.

\subsection{Hardware Experiments}
\label{ssec:hardware}

\begin{figure}[t]
    \centering
    \includegraphics[width=0.85\columnwidth]{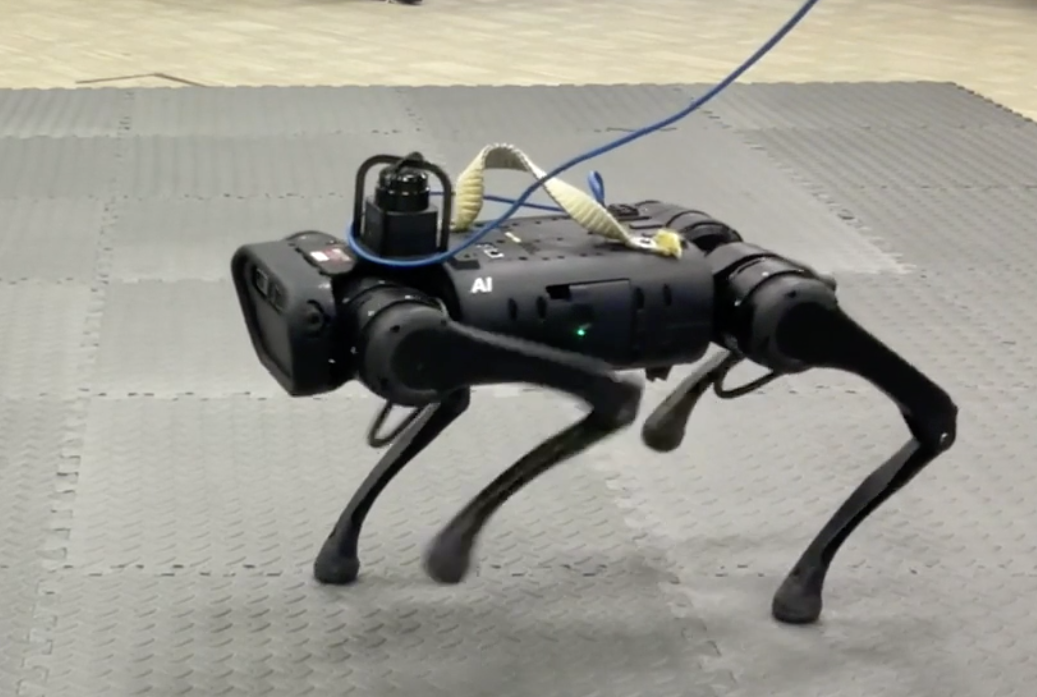} 
    \caption{Hardware experiments with the Unitree A1.
    }
    \label{fig:hardware_screenshot}
\end{figure}
Fig.~\ref{fig:hardware_screenshot} shows the Unitree A1 robot, which was connected to a desktop computer via Ethernet cable. 
The desktop computer processed sensor measurement in real time, computed torque commands, and sent the torque commands to the robot's joints.
The controller on the desktop computer used the state estimate provided by the proposed IMM-KF.

\begin{figure}[t]
    \centering
    \includegraphics[width=0.925\columnwidth]{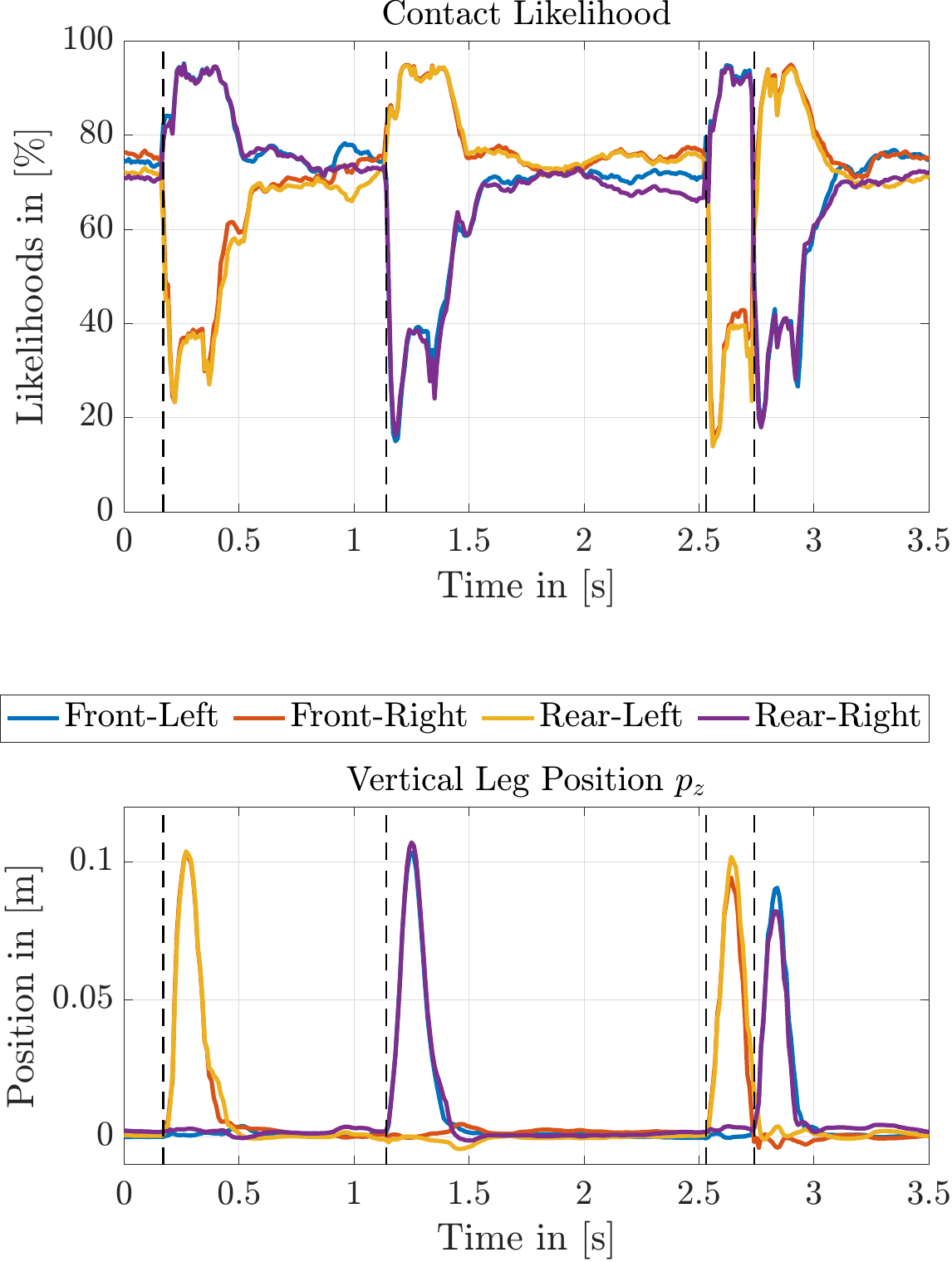} 
    \caption{Hardware experiment results.
    Top: Contact likelihoods of the four feet.
    Bottom: Feet positions calculated using forward kinematics.
    }
    \label{fig:contact_likelihoods}
\end{figure}

Fig.~\ref{fig:contact_likelihoods} shows the contact likelihoods, ${\rm Pr}(\delta_i=1)$ computed as in~\eqref{eq:calc_contact_prob} for the four legs.
It shows the robot taking four steps, which can be identified by the vertical foot positions in the bottom plot, which were computed using forward kinematics.
First, the front-right and rear-left legs swing together around 0.2s.
Second, the front-left and rear-right legs swing together around 1.2s.
At 2.5s, two trotting motions are executed back-to-back.
During a swing, the contact likelihoods of the swinging legs are below 40\%, whereas the likelihoods of the stance legs are around 95\%.
When all four legs are in contact with the ground, all contact likelihoods are around 75\%. 
The reasons for higher likelihoods of the stance legs during a swing execution are (i) higher contact forces associated with the two stance legs compared to a situation with all four legs being in contact with the ground and (ii) the physics-based biasing of probabilities of swinging legs as described in Section~\ref{ssec:force_positive}.
 
\begin{figure}[t]
    \centering
    \includegraphics[width=0.95\columnwidth]{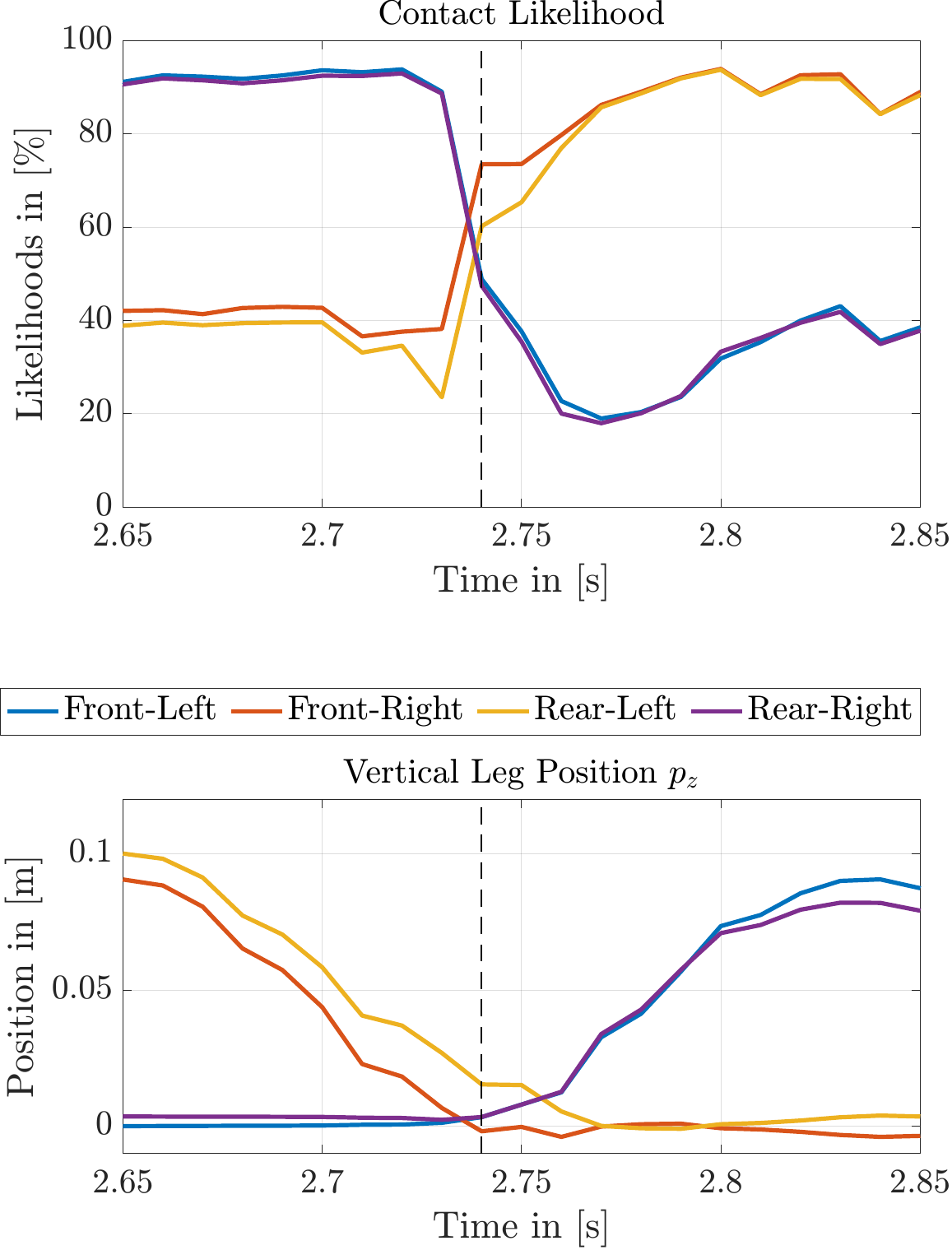} 
    \caption{Early contact detection in hardware experiment.
    Top: Contact likelihoods.
    Bottom: Feet positions calculated using forward kinematics.
    }
    \label{fig:contact_likelihoods_focus}
\end{figure}

\begin{figure}[t]
    \centering
    \includegraphics[width=0.95\columnwidth]{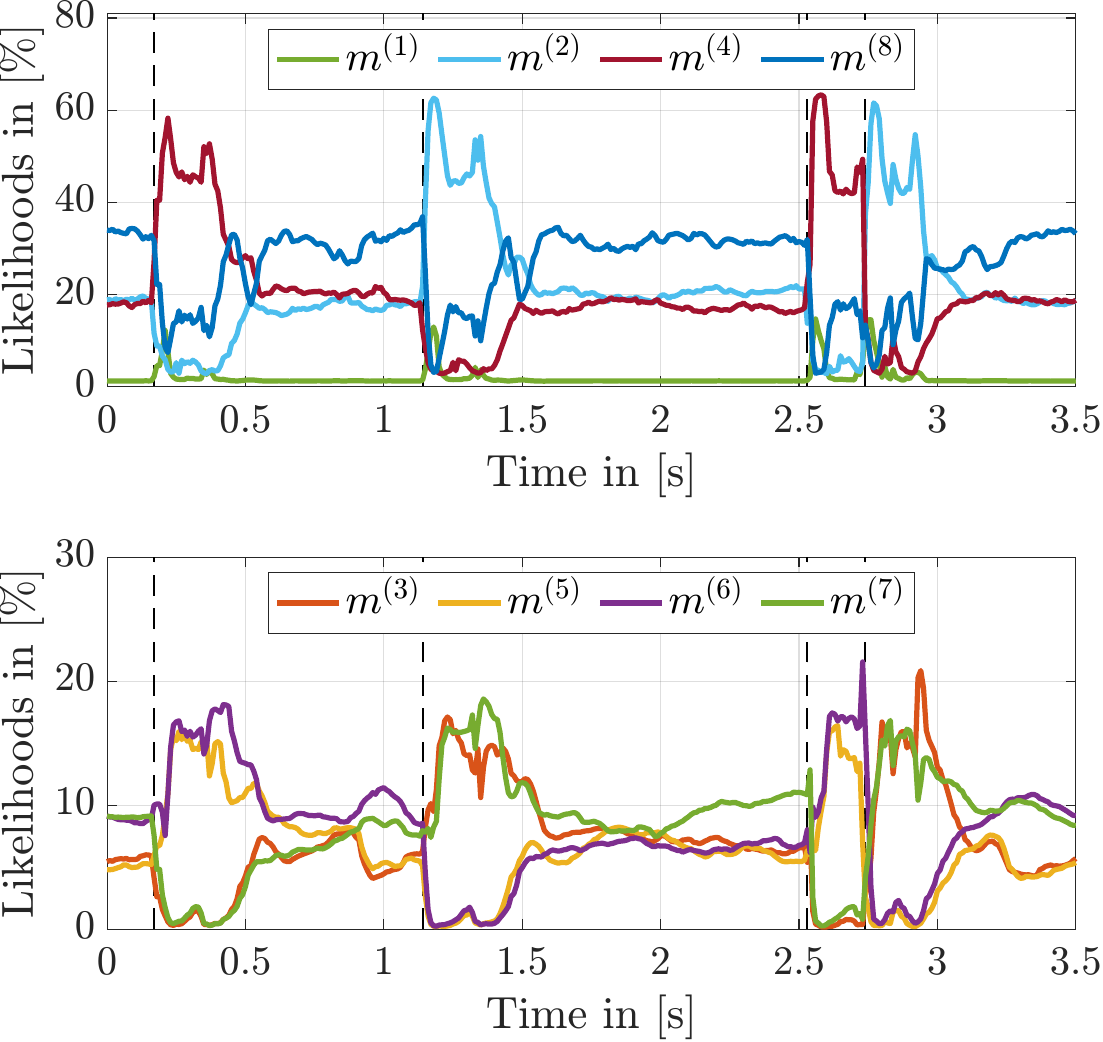} 
    \caption{Likelihoods of modes.
    Top: Mode $m^{(1)}$ (no feet in contact), mode $m^{(8)}$ (four feet in contact), and trot modes $m^{(2)}$, $m^{(4)}$ (two feet in contact).
    Bottom: Modes $m^{(3)}$, $m^{(5)}$, $m^{(6)}$, $m^{(7)}$ (three feet in contact).
    }
    \label{fig:mode_likelihoods}
\end{figure}

Fig.~\ref{fig:contact_likelihoods_focus} highlights a moment in time with an early contact.
Here, the front-right foot is touching down earlier than the rear-left foot by around 2cm or around 20ms, which can best be seen by the vertical position plot.
Consequently, the contact likelihood associated with the front-right leg increase faster than that of the rear-left leg.
Fig.~\ref{fig:mode_likelihoods} illustrates the mode likelihoods, $\mu^{(k)}$, which are used within the IMM-KF.
It shows that mode $m^{(6)}$ peaks briefly at around 2.7--2.75s identifying correctly that the front-right foot has touched down early.
It can be seen that contacts are detected using the proposed IMM-KF in less than 20ms, i.e., four filter recursions at an execution rate of 200Hz.

\subsection{Computation Times}
The model-based nature of the contact detection and state estimation makes the algorithm real-time feasible.
In particular, computation times average 2.1ms (mean), 2.0ms (median), and showed a maximum computation time of 3.5ms, which is below the sampling frequency of 5ms (200Hz).
Such high frequencies are required for contact detection in order to quickly identify feet that have touched down.
The algorithm was executed on a desktop computer using ROS in Python 2.7 with AMD 5950X, DDR4 2933 32GB RAM, and NVIDIA GeForce GTX TITAN X.

\section{Conclusion}
\label{sec:conclusion}
This paper proposed a model-based algorithm for simultaneous contact detection and state estimation. 
The algorithm used an interacting multiple-model Kalman filter, where each filter model relates to a configuration of feet being in contact with the ground.
The algorithm used IMU measurements, i.e., Euler angles, rotations, and accelerations, as well as motor measurements, i.e., joint angles, joint angular velocities, and joint torques, all of which are commonly available on robotic platforms such as a legged robot.
The interacting multiple-model Kalman filter used these sensor measurements for both estimating the robot's state and detecting which feet are in contact with the ground.
Simulation results using the high-fidelity environment Gazebo showed an improved accuracy of state estimation by a factor of 2.5 compared to a Kalman filter without simultaneous contact detection.
Further, hardware experiments showed that the algorithm is able to estimate the state accurately and in real time.
In particular, it showed that contacts can correctly be identified with only a few Kalman filter recursions in less than 20ms.

\bibliography{main.bib}
\bibliographystyle{ieeetr}

\end{document}